# Enhanced Wavelet Scattering Network for image inpainting detection

**Barglazan Adrian-Alin** adrian.barglazan@ulbsibiu.ro

*Department of Computers and Electrical Engineering, Faculty of Engineering, University "Lucian Blaga", Sibiu, Romania*

**Brad Remus** remus.brad@ulbsibiu.ro

*Department of Computers and Electrical Engineering, Faculty of Engineering, University "Lucian Blaga", Sibiu, Romania*

**Abstract**: The rapid advancement of image inpainting tools, especially those aimed at removing artifacts, has made digital image manipulation alarmingly accessible. This paper proposes several innovative ideas for detecting inpainting forgeries based on low level noise analysis by combining Dual-Tree Complex Wavelet Transform (DT-CWT) for feature extraction with convolutional neural networks (CNN) for forged area detection and localization, and lastly by employing an innovative combination of texture segmentation with noise variance estimations. The DT-CWT offers significant advantages due to its shift-invariance, enhancing its robustness against subtle manipulations during the inpainting process. Furthermore, its directional selectivity allows for the detection of subtle artifacts introduced by inpainting within specific frequency bands and orientations. Various neural network architectures were evaluated and proposed. Lastly, we propose a fusion detection module that combines texture analysis with noise variance estimation to give the forged area. Also, to address the limitations of existing inpainting datasets, particularly their lack of clear separation between inpainted regions and removed objects—which can inadvertently favor detection—we introduced a new dataset named the Real Inpainting Detection Dataset. Our approach was benchmarked against state-of-the-art methods and demonstrated superior performance over all cited alternatives. The training code (with pretrained model weights) as long as the dataset will be available at https://github.com/jmaba/Deep-dual-tree-complex-neural-network-for-image-inpainting-detection

Keyword:—computer vision, computer forensic, inpainting detection

# 1  Introduction

Image inpainting, the process of reconstructing lost or deteriorated parts of an image, is essential in fields such as image editing and restoration. Although inpainting methods have advanced significantly – as it can be observed in some through reviews done in [1] or in [2], detecting the presence of inpainting remains a critical challenge [3]. The increasing sophistication of these techniques necessitates reliable tools that can identify even subtle modifications. The proliferation of image manipulation has spurred significant growth in multimedia forensics and related disciplines. Consequently, a plethora of methods and tools have emerged for detecting and locating image forgeries. As can be seen from systematic reviews of forgery detection papers [4], while this field has expanded rapidly, current research primarily concentrates on the identification of deepfakes [5]. Although some substantial advancements have been achieved in detecting image copy-move or splicing, current state-of-the-art methods for inpainting detection remain insufficient for practical use in real-world situations. This deficiency is due to several critical challenges that are still being tackled:

- limited generalizability across various image datasets
- limited realist
- the absence of a comprehensive image dataset that encompasses multiple inpainting techniques
- limitations to image variations
- insufficient detection accuracy
- and the fast-paced evolution of inpainting technologies.

To address the above limitations, this paper proposes a novel inpainting detection framework that utilizes a wavelet scatternet based on Dual Tree Complex Wavelet in conjunction with CNN neural networks. We propose a method that combines wavelet scattering with a UNET++ architecture with an EfficientV2 like encoder. Additionally, to enhance the Intersection over Union (IoU) metric, the proposed method incorporates a fusion module at the end of the neural network (see Figure 1), which is based on texture segmentation and noise level analysis applied to the mask suggested by the neural network. Additionally, we propose a new dataset called the Real Inpainting Detection Dataset, designed to address the existing limitations in current datasets. In this dataset, the inpainted objects (removed objects) are based on real mask objects sourced from the Google Open Images dataset, rather than randomly generated masks that could potentially overlap with multiple objects in an image. The dataset incorporates several distinct inpainting methods, each applied to the same images, and includes a variety of backgrounds and architectural approaches (such as GANs, diffusion models, transformers, etc.) to facilitate a more accurate validation of detection method.

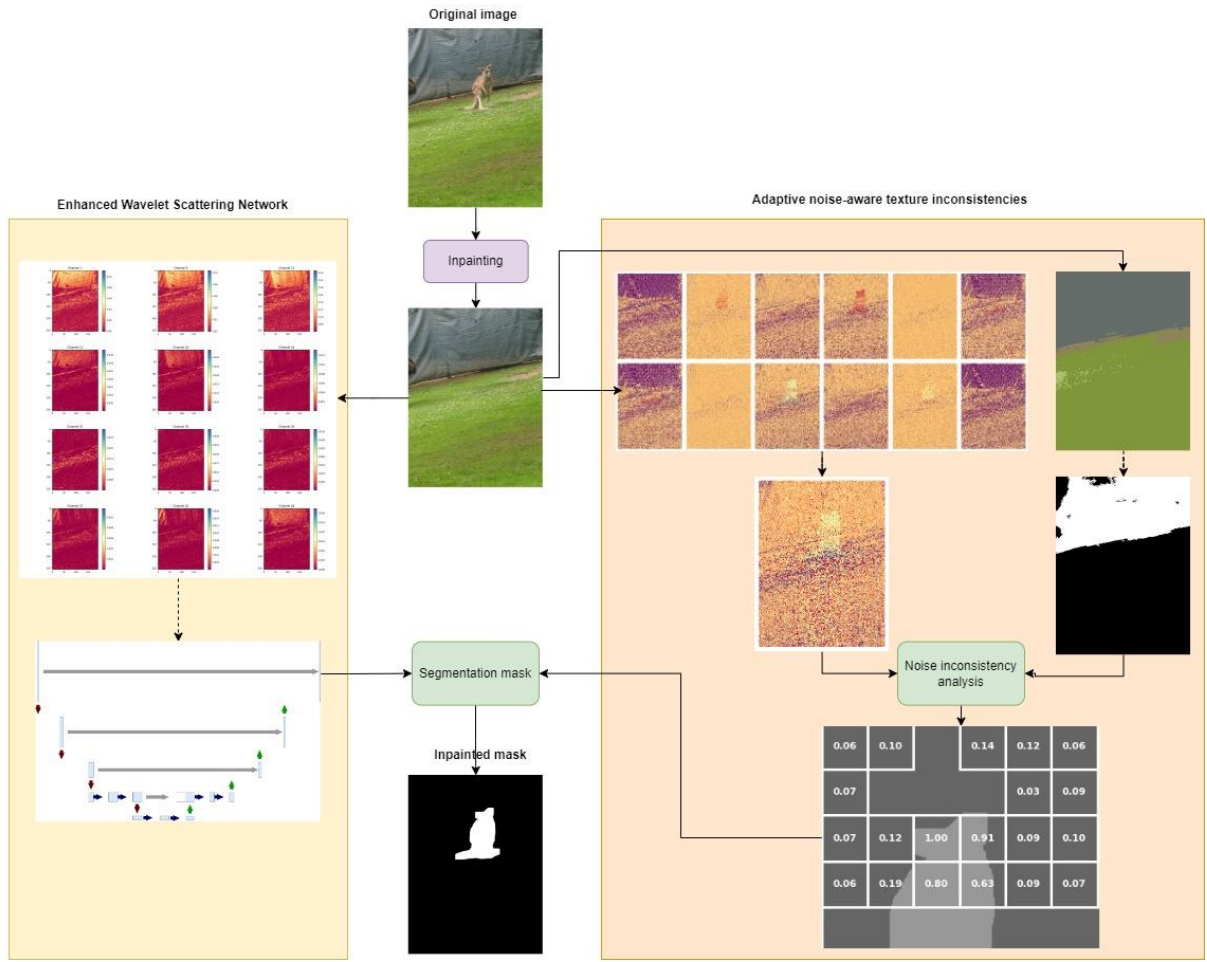

*Figure 1 Proposed Method. On the left side is the enhanced wavelet scattering network module. On the right side is the adaptive noise-aware texture inconsistencies module*

# 2  Related Work

Image forensics is a discipline dedicated to the detection and analysis of digital image manipulation. It involves techniques and algorithms to determine the authenticity, origin, and integrity of images. The primary goal is to expose forgeries, identify image sources, and provide evidence in legal investigations. Based on the classification done by the authors [6] or [7], image forgery detection can be classified based on traces: either the forgery operation specific traces like copy-move, splicing, inpainting or by optical camera traces (like blur, noise, chromatic aberration etc.). From a detection standpoint, inpainting forgery detection has not been extensively examined. It has traditionally been regarded as a subset of copy-move forgery detection. Copy-move forgery detection (CMFD) is a well-established digital image forensic technique known for its capability to detect altered regions in multimedia content. Researchers have developed numerous CMFD algorithms, drawing on either traditional digital image processing (DIP) and feature-based approaches or leveraging deep learning techniques. In the following sub-section, we will review CMFD methods applicable to image inpainting, as well as a few specific inpainting methods.

In block-based methods, the image gets divided into non-overlapping, overlapping, or partially overlapping blocks of equal sizes, and then some feature extraction methods are employed, and lastly the feature is sorted lexicography. For feature extraction several methods have been employed like using DCT [8] or DWT [9]. The method based on DWT applies image decomposition and then further processes only the LL part, thus ignoring high level details. Some authors even tried to combine these techniques [10]. All the above methods have their strengths, but the main disadvantage relies either on the inability to manage cases like resize, rotate of the copy object, or on the block related decisions: size, overlapping or non-overlapping, dimensionality reduction, etc. To overcome the above limitations, authors have suggested an approach based on key points extraction like SURF [11]or SIFT[12]. From an analysis performed in [13] and well-crafted dataset, the method with the best results is the one based on Zernike moments [14]. In general, so-called key-points method performance is better than block based, but the problem they lack is their inability to manage cases when real objects of the same textures are repeated naturally inside the image [15]. For older inpainting patch-based methods like the one proposed by Criminisi [16], some of the above methods yield decent results, but with newer patch-based methods and deep learning methods, the above methods lack also the inability to handle image inpainting detection.

In terms of newer, neural networks-based forgery detection methods, one of the most cited papers is [17]. Their approach is a universal forgery detection mechanism, by combining a feature extractor with an anomaly detection module. The feature extractor learns several types of patterns introduced in forgery processing operations (like blur inconsistencies, color / texture inconsistencies etc.). The problem with their approach is that newer methods like diffusion methods or transformers have hugely different artifacts introduced. Thus, it is always necessary to retrain the feature extractor with newer methods. Specifically, for inpainting detection methods, in [18] the authors proposed a neural network called IID-NET like the Mantranet one. IID-Net employs Neural Architecture Search (NAS) to automatically discover the most effective network architecture for inpainting detection. This approach allows IID-Net to find an optimal configuration tailored to the specific task of inpainting detection. Additionally, IID-Net incorporates an attention mechanism that helps the network focus on the region's most likely to be manipulated, enhancing its ability to detect subtle inpainting forgeries. To train with diverse types of inpainting methods, the authors in IID-NET also propose a new dataset, consisting of several inpainting methods applied. The problem with their dataset, is that the mask to be removed (filled in) by the inpainting methods is artificially and arbitrary created, thus forcing the inpainting methods to generate a lot of visible artifacts, thus improving the changes of the detection method. An improvement to the IID method was proposed in [19] called AFTLNet. The network aims to efficiently learn and detect traces left by inpainting operations, which are often subtle and difficult to identify through an adaptive learning framework. AFTLNet's performance is highly dependent on the quality and diversity of the training data. Their proposed dataset uses small images of 256x256 and again there is the problem with the generated mask of the object to be inpainted. In recent years, several improvements have been made by looking at this differently and relying only on a specific type of artifact – noise [20]. Noisesniffer operates by analyzing the noise variance across different regions of an image. It assumes that authentic images have a consistent noise pattern, while manipulated regions will exhibit anomalous noise characteristics. The tool extracts noise features from the image and uses them to create a noise map, which highlights potential tampered areas. Noisesniffer relies heavily on the assumption that authentic images have consistent noise patterns. However, this assumption may not hold for all images, especially those captured under varying conditions or with different devices, which can introduce natural noise inconsistencies. The method can be less effective on images that have undergone heavy compression. Compression artifacts can interfere with the noise extraction process, leading to false positives or false negatives in forgery detection [21].

To address all the limitations above mentioned, in the following paper, a novel based approached is proposed by incorporating scatternets first proposed by Mallat in [22] and the Dual-tree complex wavelet first proposed by Kingsbury in [23]. Wavelet transforms have been widely used in image processing tasks such as compression, denoising, and feature extraction. Wavelet scattering proposed by Mallat provides a robust representation of signals and images that is stable to deformations, such as translations and small rotations, making it particularly useful for tasks like classification and pattern recognition. The Dual-Tree Complex Wavelet Transform (DT-CWT), introduced by Kingsbury, offers advantages over the traditional Discrete Wavelet Transform (DWT) by providing nearly shift-invariant and directionally selective filters. These properties make DT-CWT particularly useful in detecting fine details and preserving edge information, which are crucial in inpainting detection. Using the above mentioned only two researchers have used – first one [24] in which the authors tried to use the artifacts for face forgery detection, and the latter one [25] which uses the same block-based approach but relies on the DTCWT for feature extractions.

# 3  Proposed method

Our proposed method involves using a Scatternet encoder / decoder architecture to learn high-level feature inconsistencies and generate an initial mask. To enhance the detection process, an additional module is incorporated, which takes the original image and applies color segmentation. For each segmented region, the module determines the extent of overlap with the mask produced by the Scatternet. Segments that either do not overlap or fully overlap with the mask are disregarded. However, if a segment partially overlaps with the mask, a noise distribution is calculated for that segment. Only irregularities in the noise distribution at the block level within these partially overlapping segments are marked as potential forgeries. A detailed procedure is presented in the Figure 2

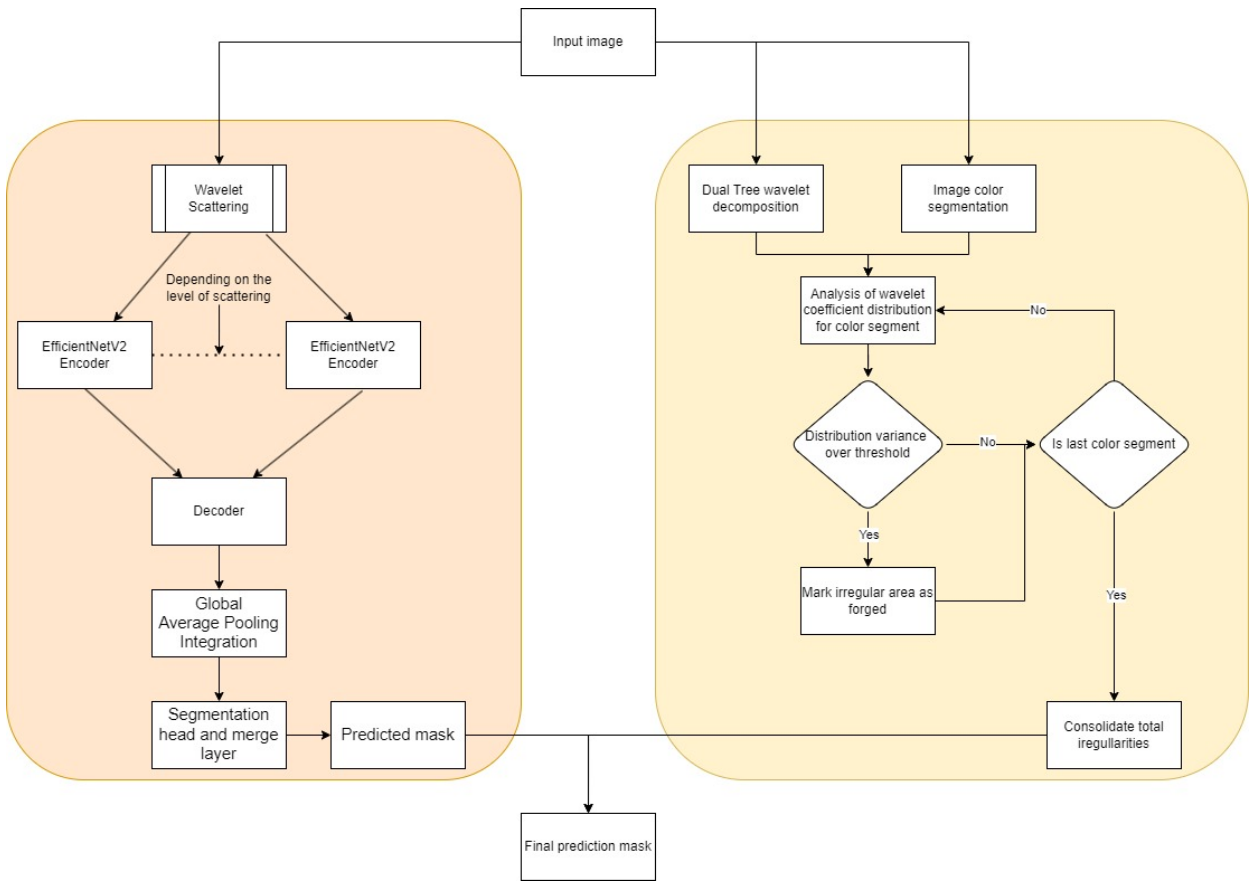

*Figure 2 Detailed description of the 2 components from our proposed method. On the left side the enhanced wavelet scattering network is presented. On the right side the noise-aware texture inconsistency module is presented*

## 3.1  Enhanced Wavelet Scattering Network

Our inspiration is drawn from recent studies in the use of complex wavelets in neural networks [26]. The first novelty of the proposed method for image inpainting detection is the use of scatternets for feature extraction – more specifically an enhanced variant of learnable scattering layer proposed in [26]. Wavelet Scatternet [22], a powerful tool for feature extraction in image forgery detection, operates on the principles of wavelet transformation and scattering. The method involves cascading wavelet transformations and modulus non-linearities to capture local frequency through wavelet coefficients, providing a comprehensive representation of image features. This approach is particularly effective in capturing various informational scales in signature images, from global components to finer details and high-frequency components. The mathematical foundation of the wavelet scatternet involves the following main operations: Wavelet Transform is a convolution with a set of wavelet filters at different scales and orientations; Modulus operation which captures the amplitude of the signal, making the representation non-linear and thus more robust to small variations in the input; Averaging (Smoothing) thus reducing the spatial resolution while maintaining essential information, enhancing the stability of the features; Nonlinearity applied to enhance discriminative information and last step is the Pooling in which features are pooled to reduce dimensionality and improve generalization. For an input image $I(x)$, the wavelet transforms at scale $2^j$ and orientation $\theta$ is given by:

$$W_j^\theta(I(x)) = I(x) * \psi_j^\theta(x)$$

where $\psi_j^\theta(x)$ is the wavelet filter. The modulus and averaging operations are then applied to form the scatternet representation:

$$S_j^\theta(I(x)) = |W_j^\theta(I(x))| * \phi_j(x)$$

where $\phi_j(x)$ is a low-pass filter. After applying the wavelet transform, the modulus operation is performed to introduce non-linearity:

$$M_j^\theta(I(x)) = |W_j^\theta(I(x))|$$

This operation ensures that the magnitude of the wavelet coefficients is considered, which helps emphasize texture discontinuities and anomalies that may arise from forgery. Non-linearity plays a crucial role in capturing complex interactions between different frequency components. To achieve translation invariance, the modulus of the wavelet coefficients is smoothed by applying a low-pass filter, typically a Gaussian filter:

$$S_j^\theta(I(x)) = M_j^\theta(I(x)) * \phi_j(x)$$

where $\phi_j(x)$ is the low-pass filter. This step ensures that small translations in the image do not significantly alter the feature representation, providing robustness to misaligned forged regions. The averaging step provides translation invariance, ensuring that small shifts or displacements in forged regions do not result in substantial changes in the scatternet representation. Mathematically, for small translations $\delta$, we have - this robustness is crucial in cases where forged regions may be slightly misaligned:

$$S_j^\theta(I(x + \delta)) \approx S_j^\theta(I(x))$$

A special type of wavelet scattering was proposed by the authors in [27]. The Dual-Tree Complex Wavelet Transform (DTCWT) extends the traditional Discrete Wavelet Transform (DWT) by using complex-valued wavelets. It involves two parallel wavelet trees (real and imaginary) that form a complex representation. The complex wavelet coefficients are formed as:

$$W_{complex}(I(x)) = W_{real}(I(x)) + iW_{imag}(I(x))$$

The magnitude of these coefficients is used to capture texture features:

$$|W_{complex}(I(x))| = \sqrt{W_{real}(I(x))^2 + W_{imag}(I(x))^2}$$

The main benefits of using DTCWT for capturing texture patterns can be summarized: shift invariance - approximately shift-invariant, making it robust for texture recognition, reduced redundancy - more efficient representation compared to other complex wavelet transforms and speed as the results presented further in paper will demonstrate it. The use of scatternets for image forgery detection has not been tried until now before, there are only a couple of papers that use DTCWT [24] for deep fake detection or copy move approach [28]. The authors in [28] they use the Dual-Tree only for feature extraction and their interest is only in the low level subband, thus ignoring valuable information is from other subbands

Our proposed encoder architecture presents several key differences from the approach outlined in [26] – see Figure 3. Firstly, the low-level band information is excluded, as our network is specifically designed to identify inconsistencies within higher components of the wavelet subbands decomposition. This focus allows the model to concentrate on the high-frequency components that are more likely to reveal subtle manipulations or discrepancies. For each individual channel—given that $n$ levels of decomposition yield a total of $6 \times n$ channels per input channel—an EfficientNetV2-like encoder architecture in [29] is employed. The choice of EfficientNetV2 is driven by its efficient scaling and superior performance in extracting meaningful features while maintaining computational efficiency. This architecture's ability to balance depth, width, and resolution ensures that the encoded features are both rich and computationally feasible. The importance of the encoders design is crucial because they transform the input into a hierarchy of feature maps, capturing fine to coarse details from each channel. This is critical for detecting subtle manipulations like inpainting, where local inconsistencies may vary across different orientations. The multi-encoder setup allows the model to effectively capture a broad range of structural and textural information, as each encoder specializes in a different aspect of the wavelet coefficients. This diversity of features is essential for identifying alterations that standard CNN architectures may overlook.

Subsequently, the outputs from each encoded channel are integrated into a decoder, structured similarly to a UNet++ architecture [30]. The decoding process within the model leverages the U-Net++ architecture, which employs densely connected skip pathways that connect intermediate encoder outputs to corresponding stages within the decoder. This design significantly enhances the flow of information and facilitates feature reuse across different levels of the network, thereby improving the model's ability to capture both local and global context. The U-Net++ decoder comprises nested decoding blocks, beginning with the deepest feature maps, characterized by low spatial resolution and high semantic content. Each block in the decoder progressively up-samples the feature maps while integrating information from the corresponding encoder stages via skip connections. This integration of multi-scale features allows the network to refine its segmentation outputs, enabling it to accurately delineate inpainted regions. The nested structure of the U-Net++ decoder provides a robust mechanism for capturing fine details and contextual information simultaneously, which is particularly crucial for distinguishing natural textures from artificially inpainted areas. By combining dense connectivity with multi-scale feature extraction, the decoder effectively reconstructs a detailed understanding of the image, facilitating the detection of subtle alterations. This architectural approach ensures that both high-level semantic features and low-level spatial features contribute to the final segmentation output, enhancing the model's sensitivity to inconsistencies indicative of inpainting.

The final stage of the model involves a segmentation head that processes the outputs from the U-Net++ decoder. This component consists of a convolutional layer followed by a sigmoid activation function, which transforms the multi-channel decoder output into a single-channel segmentation mask. The sigmoid activation constrains the pixel values to a range between 0 and 1, representing the probability of each pixel being part of an inpainted region. This probabilistic approach to segmentation allows the model to express varying degrees of confidence regarding the likelihood of inpainting at each pixel location. The generated segmentation mask serves as a direct visual representation of the model's predictions, highlighting areas that are deemed to be altered. This output is crucial for inpainting detection, as it provides a pixel-level identification of manipulated regions, allowing for a detailed and interpretable assessment of the image's authenticity. The ability to produce precise segmentation masks is instrumental in forensic applications, where identifying even subtle inpainting artifacts can be critical.

The model's output is a probabilistic segmentation mask that identifies areas within the image suspected of being inpainted. Each pixel value in the mask represents the model's confidence in whether that specific pixel has been artificially altered. By integrating multi-scale feature extraction with dense connections throughout the decoding process, the model can detect nuanced artifacts commonly introduced by inpainting techniques, such as unnatural transitions, blurred edges, or repetitive patterns that deviate from the original texture of the image. The incorporation of DTCWT coefficients enhances the model's ability to detect structural anomalies, providing it with a unique sensitivity to frequency-domain features that are often indicative of inpainting. This approach allows the model to effectively distinguish between naturally occurring textures and those that have been artificially manipulated, offering a sophisticated and reliable tool for detecting image alterations. By leveraging advanced convolutional networks combined with wavelet-based feature extraction, the model demonstrates a high degree of accuracy in identifying inpainted regions, underscoring its potential applicability in forensic and image authentication contexts.

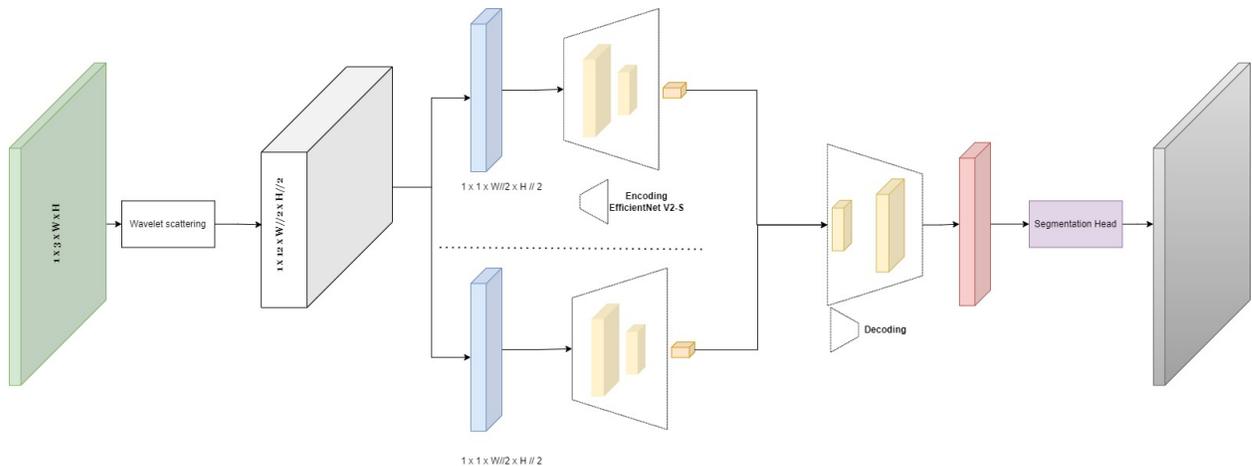

*Figure 3 Enhanced Wavelet Scattering composed of a wavelet scattering block followed by n EfficienetV2-S encoders, combined in one decoder and with a segmentation head*

## 3.2 Adaptive noise-aware texture inconsistencies

To improve the accuracy of the proposed network architecture above, extract module is added to increase the accuracy. Our proposed method for inpainting detection assumes that when inpainting is applied to an image, the removed object is typically enclosed within a larger area of similar texture (see Figure 4**Error! Reference source not found.**). This assumption remains valid even when the object is located at the boundary between two distinct textured regions, as these regions can be analyzed independently (see Figure 5). The method also relies on the observation that inpainted regions do not possess the same high-level features as the surrounding areas with similar textures.

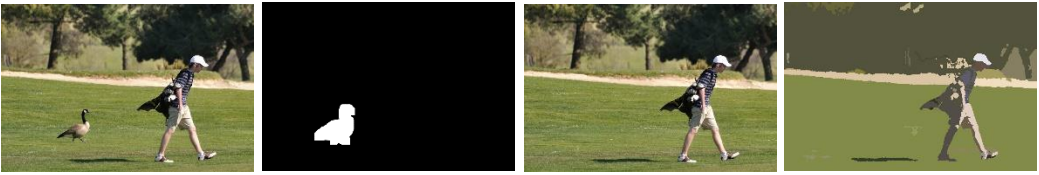

*Figure 4 Original image on the left. Mask image - object to be removed on 2$^{rd}$ column. Inpainted image on 3$^{rd}$ column. Texture segmentation on the last column*

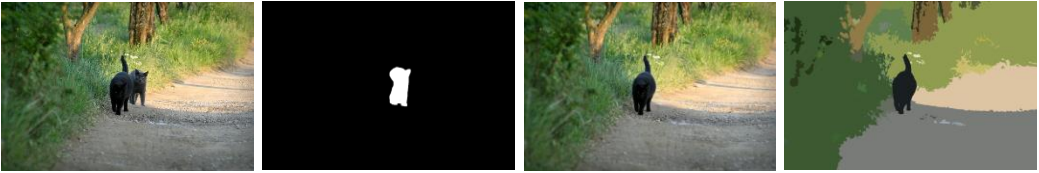

*Figure 5 Original image on the left. Mask image - object to be removed on 2$^{nd}$ column. It can be observed that the removed object is at the border between 3 different textured area – grass and road. Inpainted image on the 3$^{rd}$ column. Texture segmentation on the last column*

The proposed module called Adaptive Noise-Aware texture inconsistencies is performed via Hierarchical Feature Selection [31] and DTCWT combined with noise level estimation. Our paper introduces an advanced methodology for detecting inconsistencies in segmented color regions of an image. The process begins with Hierarchical Feature Selection for color segmentation, ensuring precise identification of regions based on color and texture characteristics. For each segmented color region, the module extracts the Dual-Tree Complex Wavelet Transform (DTCWT) coefficients, focusing on a one-level decomposition that results in six complex bands. These coefficients are specifically extracted from the positions indicated by the segmented color regions (see below image on the first row the original image and the mask of the object to be removed, while on the second row the inpainted image and the texture segmentation obtained for the inpainted image).

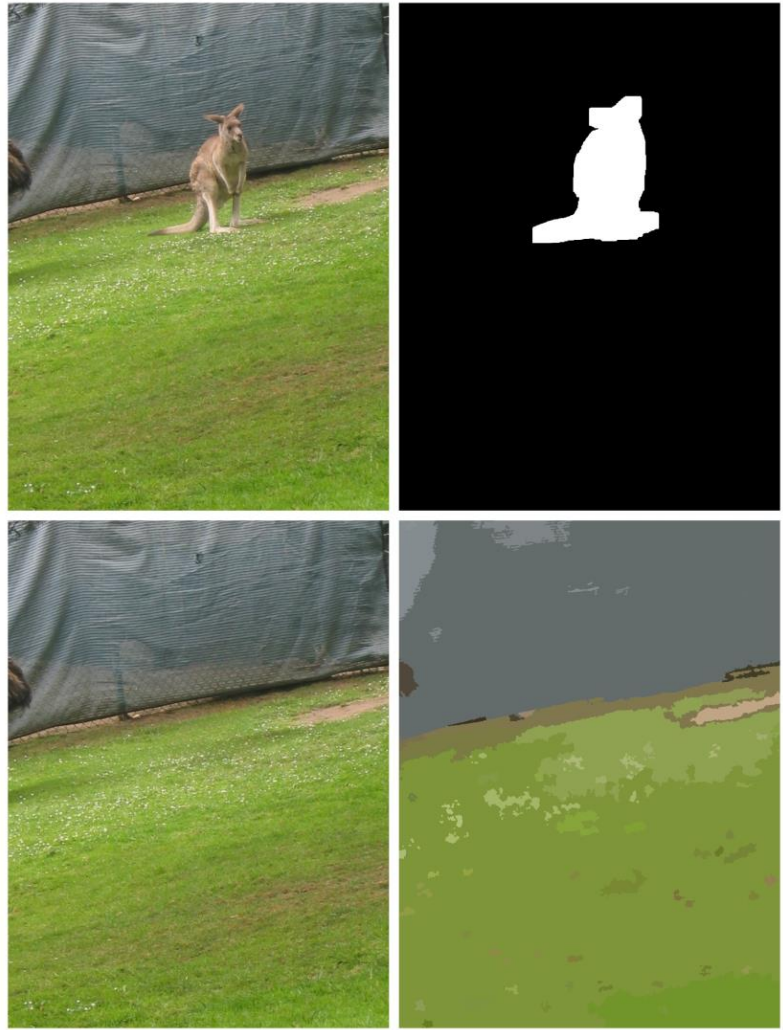

*Figure 6 On the first row: first image from left the original image, on the right the mask of the object to be removed (inpainted. On the second row: on the first column the inpainted image, and the second column the texture segmentation applied to the inpainted*

Following the segmentation, the module divides each region into smaller patches, where noise estimation is performed. The noise estimation technique is based on the method outlined in [32], which effectively estimates noise levels by analyzing image gradients, demonstrating both accuracy and efficiency across diverse image types. This method is particularly beneficial for tasks necessitating precise noise quantification. In our approach, we extend this concept by applying noise estimation to individual patches within color-segmented regions, rather than the entire image. This localized analysis allows us to detect discrepancies in noise levels within each segment. Specifically, by calculating noise levels for each patch and identifying regions with high noise variance, we flag these as suspect areas (see below image – output of applying DTCWT to the inpainted image - Figure 7 ). In Figure 8, the first image represents the output of texture segmentation, the second image represents the first segment / area to be analyzed (our proposed method analyzes all textures with a size threshold given), while the last image represents the patches together with their noise estimation from DTCWT. This targeted method enables a more precise detection of noise inconsistencies, which are often indicative of image forgery or tampering. A more zoom result version can be seen in Figure 9

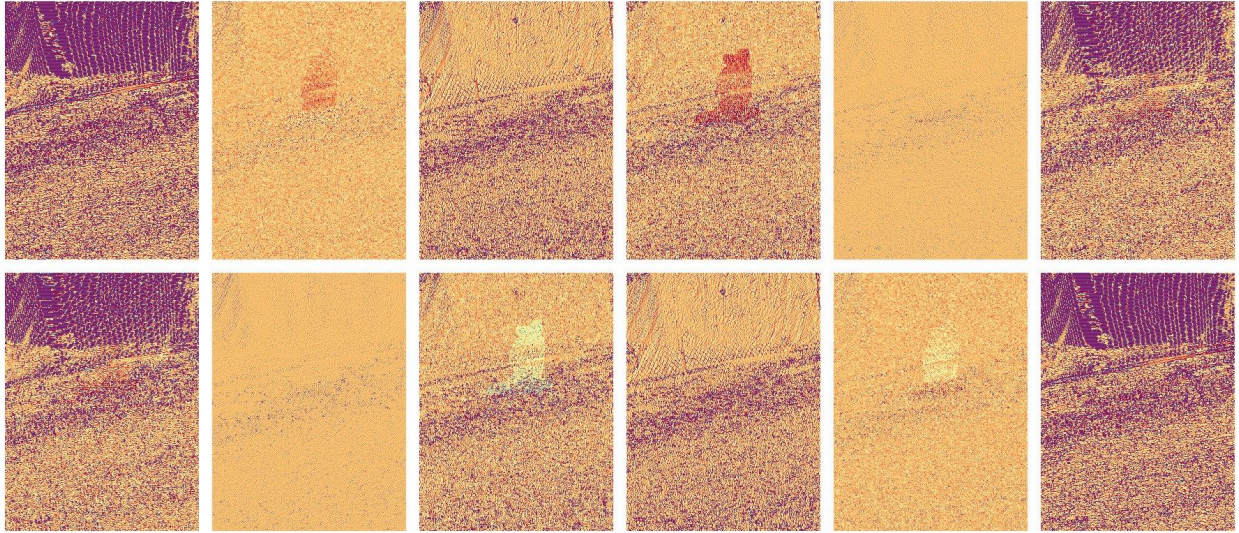

*Figure 7 Output of applying DTCWT to the inpainted image*

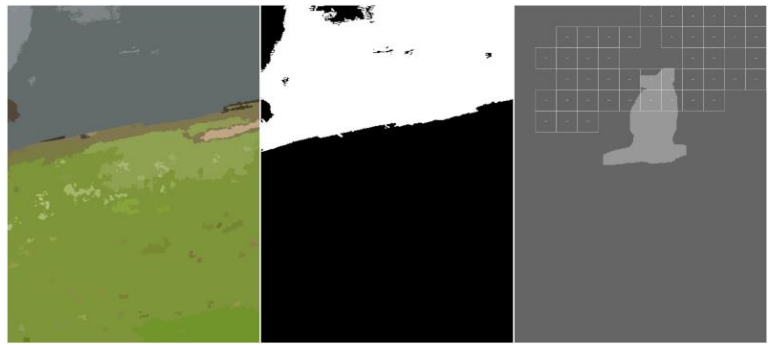

*Figure 8 First image represents the texture segmentation. The second image represents the current mask / segment to be processed. Last image display on the current segmented mask, the patch noise level estimation using DTCWT*

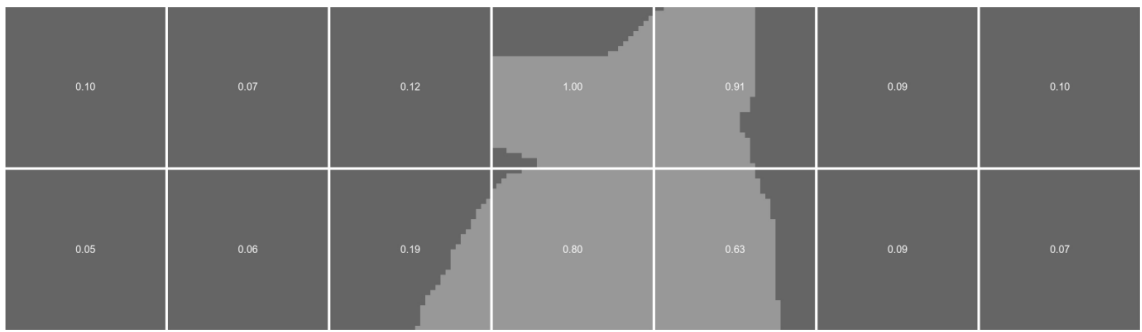

*Figure 9 Zoom in result displaying Inconsistencies in patch mean for second band of real values of the DTCWT*

The last step in the module's process involves comparing the suspected regions with a potential mask generated from a neural network. If the areas of high noise variance overlap with the mask, these regions are conclusively marked as forged, ensuring a robust detection of tampered areas. Conversely, regions without noise inconsistency are disregarded, thereby refining the final mask to only include areas with confirmed inconsistencies. This method enhances the reliability of color segmentation and noise detection in complex image analysis tasks.

# 4 Results

## 4.1 Real Inpainting Detection Dataset

Building on our previous research in [3], it is evident that existing inpainting detection datasets are insufficient for real-world applications. These datasets often lack critical attributes necessary for accurately identifying inpainted regions (e.g., removed objects). Even in recent works like [19] or [18], the proposed datasets suffer from inconsistencies where the inpainted regions do not align well with the original image content. To improve the effectiveness of inpainting methods, we propose focusing on clearly defined objects or regions. For this reason, we employ Google Open V7 dataset [33] and for the segmented objects we rely on the [34].

To evaluate the effectiveness of our proposed method, we applied it alongside several inpainting techniques. Specifically, for each image in the Google Open Images dataset, we selected a segmented object and applied three different inpainting methods. This process resulted in a total of 12,000 forged images from an initial input of 4,000 images. All original images, masks and inpainted mages are uploaded to https://github.com/jmaba/Deep-dual-tree-complex-neural-network-for-image-inpainting-detection. The inpainted methods are taken to be from different approaches like Fourier based [35] or newer methods based on transformers like [36] and [37].

## 4.2 Experimental setup

During the training process, 6k of the images were randomly selected from each inpainting method. A separate set of 1,500 images was used for validation, while the rest were used for testing. To ensure a robust evaluation of the model, we implemented a k-fold cross-validation strategy, allowing us to assess the model's performance across various data subsets. The model was trained for 30 epochs, providing enough iterations for optimization convergence while minimizing the risk of overfitting. This method allowed for a thorough assessment of the model's ability to generalize to unseen data. The AdamW optimization algorithm was employed with a learning rate set to 1e-5, while the weight decay was set to 1e-4.

## 4.3 Implementation details

### 4.3.1 Enhanced wavelet scattering network module

For the wavelet scattering network, two methods have been employed: first option is the standard Morlet wavelet scattering proposed by Mallat and improved in [38], and the second option is the Dual-Tree complex wavelet scattering in[26]. For the first wavelet scattering, the transform used in this study applies two successive wavelet transforms, each followed by a modulus non-linearity, utilizing eight different angles for the wavelet transform. For the latter, which is ten times faster than the former, we use a second order wavelet scattering network. To determine which wavelet scattering yields the best results, the same training and validation was done for 20 epochs. From the table below it can be observed that indeed the Morlet based approach gives better results but not with a significant difference compared to the Cotter method and comparing training / testing time the Cotter approach is ten times faster. Based on these results, the used scattering method is the one proposed by Cotter.

For the EfficientNetV2 model architecture, the EfficientNetV2-S encoder is applied to each wavelet scattering channel band. The decoder utilizes an architecture like U-Net++ for each individual band, which is then followed by a segmentation head. A unique feature of the proposed model is the incorporation of global average pooling to extract global contextual information from the decoder output. This global context is processed through a fully connected layer to produce a feature that summarizes the entire image's content. The global feature is then merged with the high-resolution output from the decoder via a custom convolutional layer. This fusion of local and global information enhances the model's ability to detect inpainting across different scales.

### 4.3.2 Adaptive noise-aware texture module

The segmentation result from the network is passed into a noise-aware texture module. Initially, color segmentation is performed on the input image using the method from [31] with SLIC set to 32. Segmented regions smaller than 0.1 of the original image size are discarded. The segmented image is then patched using an eight-by-eight grid. For noise variance analysis, the original six subbands (magnitude values) are used. If the standard deviation between patches exceeds a certain threshold, those patches are flagged as suspicious. Only areas where suspicious regions overlap with the segmentation results from the previous network module are marked as forged.

## 4.4 State-of-the-art analysis comparison

Like previous works, we evaluate performance using IoU, F1, Precision, Recall, and Accuracy at pixel level. Results were reported using the default threshold of 0.5. For image-level analysis, we focus on balanced accuracy, which considers both false positives and false negatives, with the threshold set to 0.5. The proposed method was evaluated using an extended version of the publicly available inpainting forgery detection dataset. The results demonstrate the method's effectiveness in detecting tampered regions across a wider variety of image content and inpainting techniques. For demonstration purposes, we selected several images, including the originals, masks, and inpainting outcomes. We applied the detection methods proposed in [18], referred to as IID; [39], referred to as PSCCNET;[40], referred to as FOCAL and TruFor referred as [41]. For all models, the pretrained networks available were used. For [19] (a variant of [18]), we have tried training it on our dataset, but it yields not so reliable results during training. We hypothesize that this issue arises because the network requires input images of a fixed size (256x256). Consequently, the resizing operation at the beginning of the training process may distort critical features in the images, preventing the network from effectively learning. In Figure 10 the original image, mask of the removed object and the inpainted image are presented. The inpainted image is the input to detection methods. In Figure 11 the results of inpainting detection are presented – our method with comparison to state-of-the-art methods.

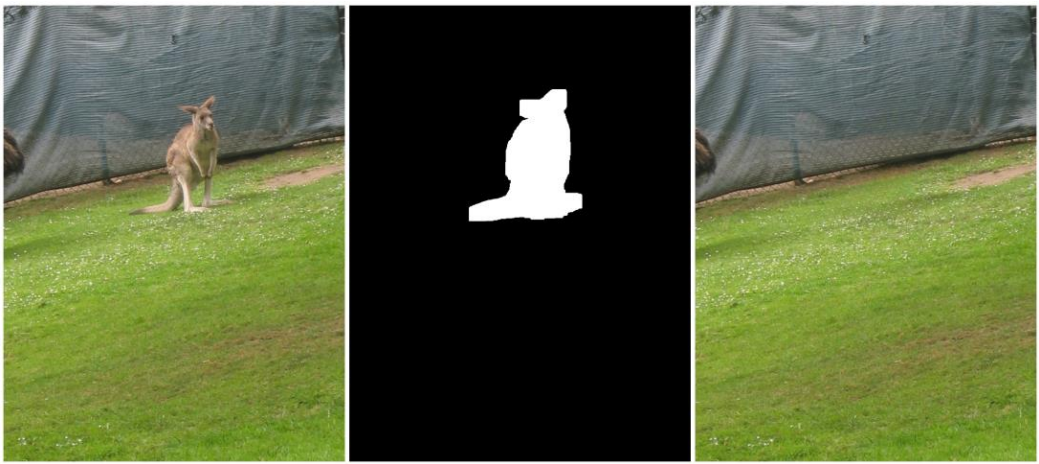

*Figure 10 Original image on the left. Mask of removed object in the middle. Inpainted image on right using LAMA*

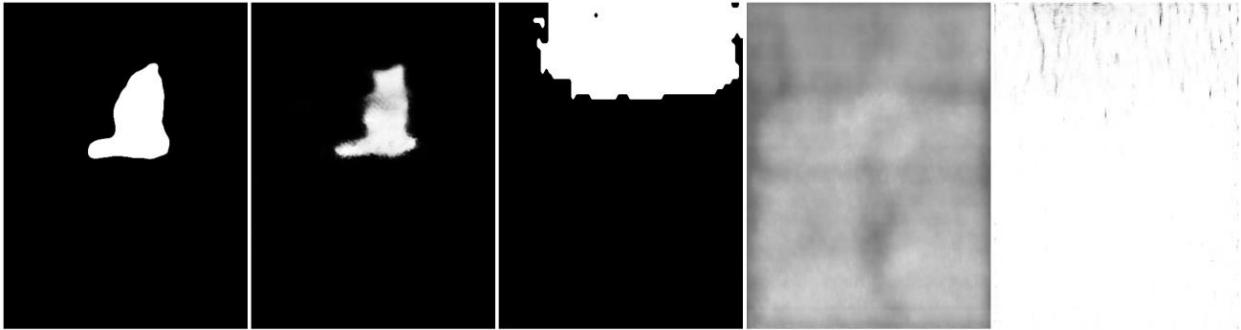

*Figure 11 Results of inpainting detection for: our method, TrueFor, FOCAL, PSCCNET, IID*

In **Error! Reference source not found.** the metrics for each detection methods for each dataset are presented.

The evaluation of various methods (FOCAL, IID, PSCCNET, TruFor) against our proposed method in the task of image inpainting detection highlights the superiority of our approach across key metrics: Accuracy, IoU (Intersection over Union), and Precision. Our proposed method consistently demonstrates the highest performance across all metrics, indicating its robustness in accurately identifying inpainted regions. Compared to other methods, it shows a marked improvement in Accuracy, reflecting superior reliability in detection. The significantly higher IoU values highlight its exceptional ability to precisely localize inpainted areas, which is crucial for detecting subtle modifications in images. Additionally, the outstanding Precision of our method—far exceeding that of FOCAL, IID, and PSCCNET—underscores its effectiveness in minimizing false positives, a critical factor for practical application. Overall, our proposed method outperforms existing approaches, establishing itself as a leading solution for image inpainting detection. Its consistent superiority across Accuracy, IoU, and Precision metrics suggests that it provides more reliable, precise, and confident detection of inpainting artifacts. These results position our method as highly effective for applications in image forensics, content verification, and quality assessment, reinforcing its potential as a state-of-the-art tool in the field.

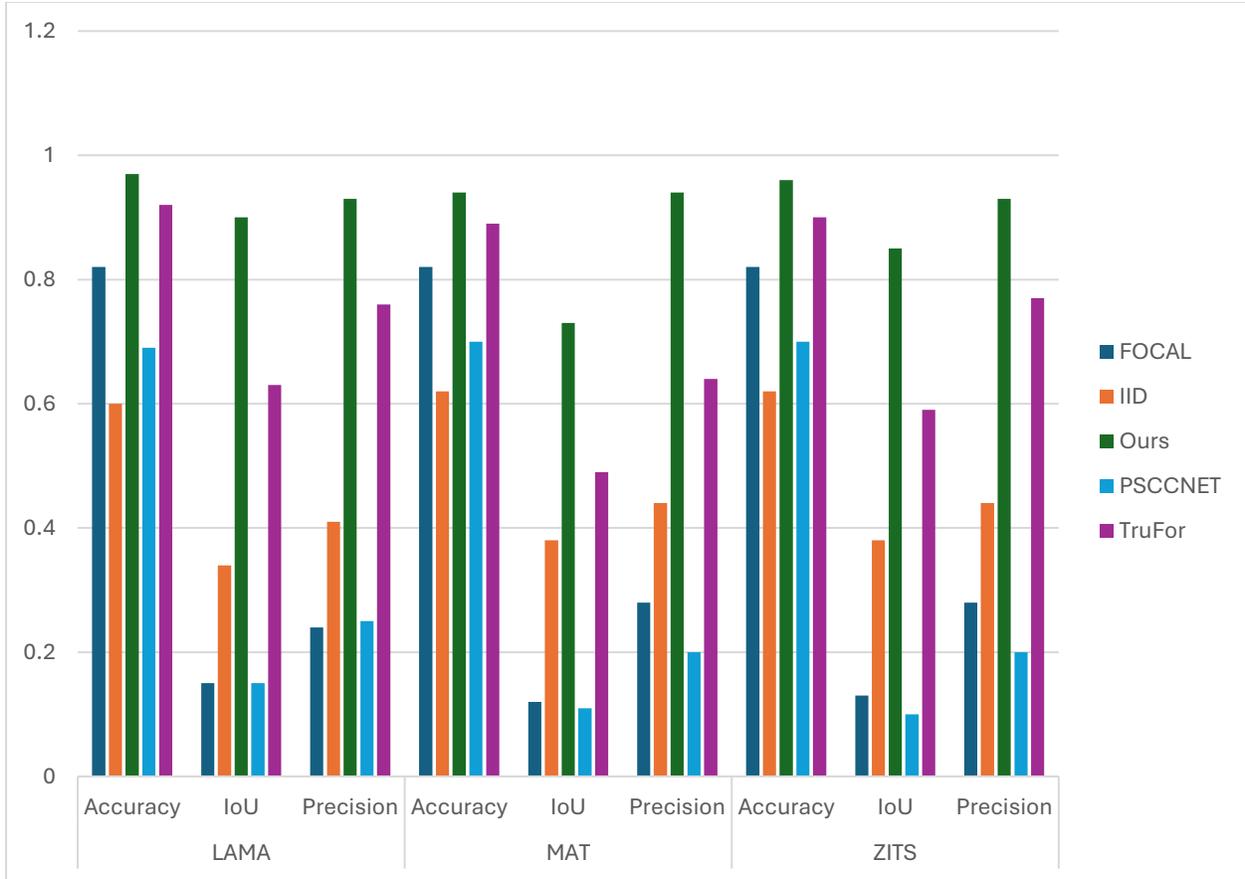

*Figure 12 Results of inpainting detection applied on 3 datasets: LAMA, MAT, ZITS. Datasets are generated with methods: LAMA, ZITS, MAT. Detection methods are our proposed method, IID, PSCCNET, FOCAL, IID and TruFor*

## 4.5 Ablation study

To evaluate the effectiveness of the proposed method, we've focused on training only on images from one dataset and testing on the others. As shown in the Table 1, the IoU was suboptimal, indicating that training exclusively on the LAMA dataset does not yield satisfactory results. This finding suggests that the network may need to be specifically trained for each category of image inpainting method. Additionally, it highlights that even in the absence of visual cues in the Dual Tree Complex Wavelet coefficients, the network still learns relevant information.

*Table 1 Results of training on one dataset and testing on the other datasets*

|  | IOU | PRECISION | ACCURACY |
| --- | --- | --- | --- |
| **TRAINED ON ALL DATASETS** | 0.82 | 0.93 | 0.95 |
| **TRAINED ON LAMA TESTED ON MAT** | 0.20 | 0.22 | 0.42 |
| **TRAINED ON LAMA TESTED ON ZITS** | 0.12 | 0.16 | 0.41 |

The IoU is significantly higher when the model is trained on all datasets (0.82) compared to when it is trained on Lama alone and tested on other datasets (0.20 for MAT and 0.12 for ZITS). This stark reduction highlights the importance of diverse training data in learning a more generalized feature set that can effectively capture inpainted regions across different test sets. Precision drops sharply when the model is trained on Lama and tested on other datasets, with values of 0.22 for MAT and 0.16 for ZITS, compared to

0.93 when trained on all datasets. This indicates that the method struggles with false positives when trained on limited data, underscoring the poor transferability of features learned solely from the Lama dataset. Accuracy also drops sharply when the model is trained on Lama and tested on other datasets, with values of 0.42 for MAT and 0.41 for ZITS, compared to 0.95 when trained on all datasets. This indicates that the method struggles with false positives when trained on limited data, underscoring the poor transferability of features learned solely from the Lama dataset. The significant decline in performance metrics when the model is trained only on the Lama dataset and tested on MAT or ZITS highlights the model's overfitting to the specific characteristics of the Lama dataset. The lack of diverse training data restricts the model's ability to generalize, making it vulnerable when exposed to unseen test data with different textures, artifacts, or inpainting patterns. The results emphasize the critical need for diverse and comprehensive training data to ensure the proposed method's robustness and effectiveness across various test conditions. Training on a wider range of datasets enables the model to capture a broader spectrum of inpainting features, resulting in significantly improved IoU, Precision, and Accuracy. These findings suggest that future work should prioritize the inclusion of varied inpainting patterns and artifacts in the training phase to enhance the model's generalizability and performance across different application scenarios.

## 4.6 Post-processing impact of forgery detection

Building on the results presented above, the following sub-chapter delves deeper into the capabilities of the proposed method in detecting alterations. We focused on two common operations—image resizing and blurring. Given that the input dataset comprises images of varying sizes, we evaluated detection performance using proportional resizing. Additionally, we evaluated the impact of blurring the input image separately, and finally, we combined resizing and blurring into a single operation.

The first post-processing operation analyzed is resize – see Figure 13 and Figure 14. The images are resized to 0.7 of their initial size. As can be noticed the proposed method does not yield good results – see Table 2.

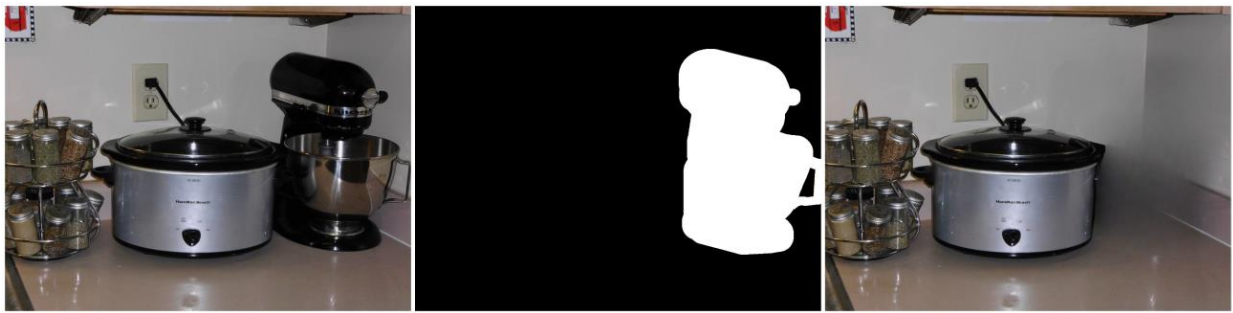

*Figure 13 On the left side the original image. On the middle the object to be removed. On the right side the inpainted image*

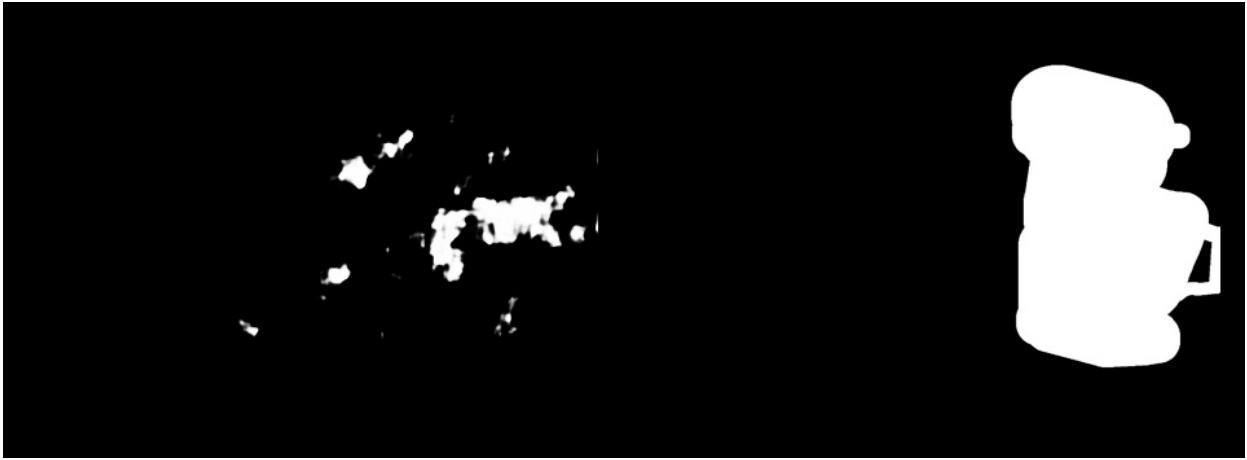

*Figure 14 After applying image resize to the inpainted image, our proposed method detects the image on the left. On the right it is the true mask.*

The IoU drops dramatically from 0.82 to 0.13 when the images are resized, indicating that the resized images severely impair the method's ability to correctly overlap detected inpainted regions with the true regions. This significant reduction suggests that the method struggles to maintain consistent detection performance when subjected to size alterations. The precision decreases from 0.93 to 0.76 after resizing, showing that the method becomes less reliable and introduces more false positives when identifying inpainted areas. Although precision remains relatively high, the drop indicates that resizing introduces noise or artifacts that affect detection quality. Accuracy also suffers, decreasing from 0.95 to 0.78. This decline reflects the method's overall reduced effectiveness in distinguishing inpainted from non-inpainted regions under resizing conditions. The primary cause of this degradation is the sensitivity of the Discrete Transform Complex Wavelet Transform (DTCWT) coefficients to resizing operations. DTCWT coefficients are critical features used by the proposed method for detecting inpainted regions. However, resizing alters the spatial frequency and orientation of these coefficients, leading to misalignments and incorrect feature extraction, ultimately impairing the detection capability. The results demonstrate that the proposed method's performance is highly susceptible to resizing operations, particularly due to the disruption of DTCWT coefficients. This highlights a significant limitation when applying the method to images that undergo resizing, emphasizing the need for robust adaptation or alternative feature extraction strategies to handle such transformations effectively. Future work should focus on enhancing the resilience of the detection algorithm to maintain performance across various post-processing operations, including resizing.

*Table 2 Resize impact analysis*

|  | IOU | PRECISION | ACCURACY |
|---|---|---|---|
| **ORIGINAL SIZE** | 0.82 | 0.93 | 0.95 |
| **RESIZED IMAGES** | 0.13 | 0.76 | 0.78 |

The second investigated post-processing operation is blurring. For this operation, we've taken the original images and applied a box blurring with a radius of 5. The visual results can be seen in Figure 15 while the overall results can be seen in Table 3

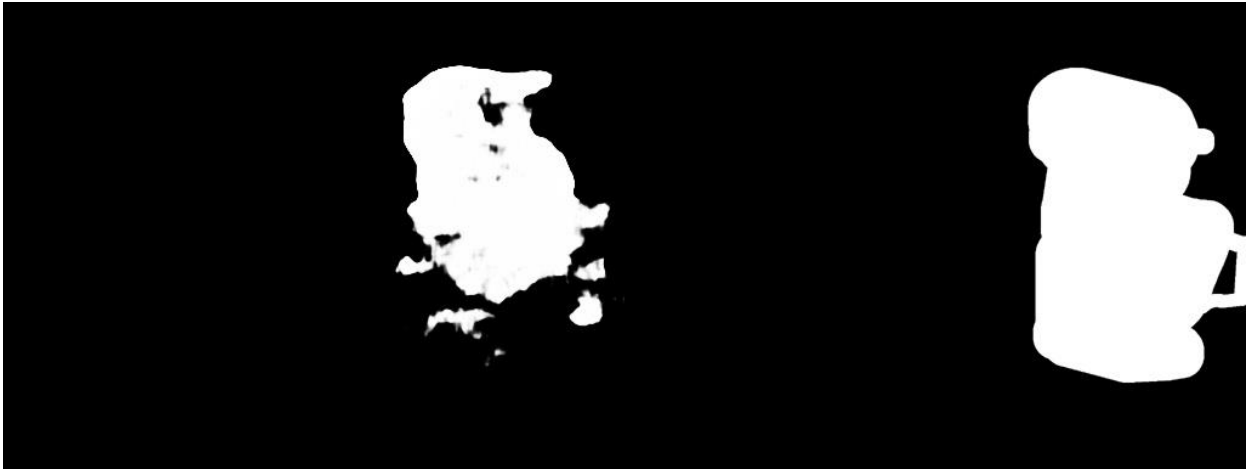

*Figure 15 On the left side the result of our proposed method after blurring the input image. On the right side the true mask*

*Table 3 Blurring impact analysis*

|  | IOU | PRECISION | ACCURACY |
|---|---|---|---|
| **ORIGINAL IMAGES** | 0.82 | 0.93 | 0.95 |
| **BLURRED IMAGES** | 0.33 | 0.92 | 0.85 |

The IoU decreases from 0.82 to 0.33 when blurring is applied, indicating a moderate reduction in the method's ability to correctly identify the overlapping areas between the detected and actual inpainted regions. Although the IoU is lower for blurred images, the method retains some effectiveness in localizing inpainted regions despite the added noise from blurring. Precision remains relatively stable, with a slight decrease from 0.93 to 0.92. This suggests that blurring does not significantly impact the method's ability to maintain a low rate of false positives. The near-constant precision indicates that the method is still confident in its positive detections, even with blurring applied. The accuracy drops from 0.95 to 0.85, showing that blurring reduces the method's overall effectiveness in correctly distinguishing inpainted regions from non-inpainted areas. The decreased accuracy points to challenges in correctly classifying the altered pixel values introduced by the blurring effect. Blurring, particularly box blurring, smooths the image by averaging pixel values within a given radius, which disrupts the edge and texture information critical for inpainting detection. The proposed method relies on detailed feature extraction, and blurring diminishes the distinctiveness of inpainted regions, making them harder to detect accurately. While the proposed method's performance degrades under blurring, the impact is less severe compared to resizing. The IoU and accuracy decline, indicating challenges in detecting precise inpainted boundaries and maintaining overall detection accuracy. However, the high precision suggests that the method remains effective at minimizing false positives despite blurring. These results highlight the method's partial robustness to blurring but also underscore the need for enhancement strategies to mitigate the effects of such image transformations. Future research should explore adaptive filtering techniques or more resilient feature extraction methods that can better withstand blurring without compromising detection quality.

# 5 Conclusions

In this paper, we present a pioneering method for inpainting detection and localization, termed Enhanced Wavelet Scattering, marking the first application of wavelet scattering networks for this task, distinct from their limited prior use in DeepFake detection with the Dual-Tree Complex Wavelet Transform (DTCWT). Our approach uniquely combines features extracted from a wavelet scattering network with a UNet++-inspired architecture and an innovative fusion module, significantly enhancing localization accuracy. The fusion module plays a pivotal role by detecting inconsistencies at the texture and color segmentation levels and analyzing high-level feature discrepancies captured by the dual-tree complex wavelet. A major contribution of this work is the introduction of a novel and extensible inpainting detection dataset, designed to address the limitations of existing datasets. This dataset offers a unique advantage by featuring the same images with objects removed and inpainted using multiple inpainting methods. This setup not only serves as a rich benchmark for evaluating detection algorithms but also allows for detailed analysis of how different inpainting techniques affect detection performance, providing deeper insights into the strengths and weaknesses of our proposed method.

Our results demonstrate that the Enhanced Wavelet Scattering approach can accurately localize inpainting manipulations across a wide range of techniques, providing a comprehensive evaluation of different inpainting methods. However, the method's effectiveness can be impacted by image post-processing operations. To address this challenge, we explored several strategies to manage network complexity, including training each wavelet subband separately and integrating the results through a fusion block, leading to promising enhancements in robustness. This study not only introduces a novel detection framework but also sets a new standard with a versatile and extensible dataset, offering valuable insights for future research in forgery and inpainting detection.

# 7 Declarations

## 7.1 Contributions

Adrian-Alin Barglazan: Conceptualization, Methodology, Investigation, Writing- Original draft preparation, Software. Remus Brad: Supervision, Validation, Writing - Review and Editing.

## 7.2 Availability of data and materials

The dataset along side training / testing code shall be uploaded in the following git repository: https://github.com/jmaba/Deep-dual-tree-complex-neural-network-for-image-inpainting-detection

## 7.3 Competing interests

The author(s) declare(s) that they have no competing interests

## 7.4 Funding

This research received no specific grant from any funding agency in the public, commercial, or not-for-profit sectors.

## 7.5 Corresponding author

Correspondence to Adrian Barglazan – adrian.barglazan@ulbsibiu.ro

## 7.6 Acknowledgements

No acknowledgements.

## 7.7 Ethics declarations

The authors declare that they have no known competing financial interests or personal relationships that could have appeared to influence the work reported in this paper.

The author confirms responsibility for data collection, analysis and interpretation of results, and manuscript preparation.

# 8 Additional image visual results

Below are some results from inpainting / detection methods. On Original image on the left and mask of the object to be removed on the right (the mask also represents the ground truth of the detection methods)Figure 16 the original image is presented on the left, while on the right the mask of the object to be removed. The mask object also represents the truth for the detection methods. On the next image - Figure 17 the inpainted images using the 3 methods are presented: on the first image the LAMA method was applied, on the second image MAT image has been applied, while on the last image the ZITS method was applied. The last image Figure 18 represents the results of the detection methods. On first column the results on the LAMA image are presented, on the second column the results on the MAT image are presented and on the third column results on the ZITS methods are presented. The rows represent the results for (in the following order): FOCAL, IID, OURS, PSCCNET, TRUFOR.

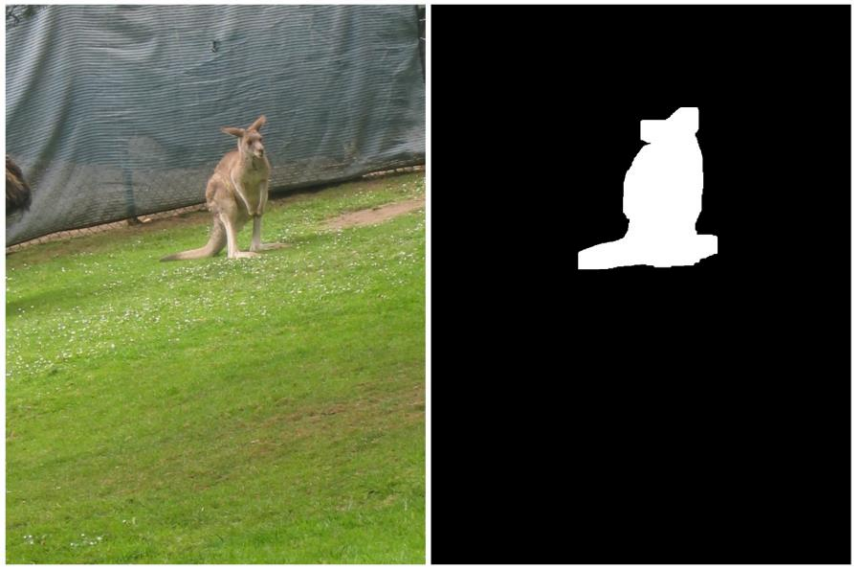

*Figure 16 Original image on the left and mask of the object to be removed on the right (the mask also represents the ground truth of the detection methods)*

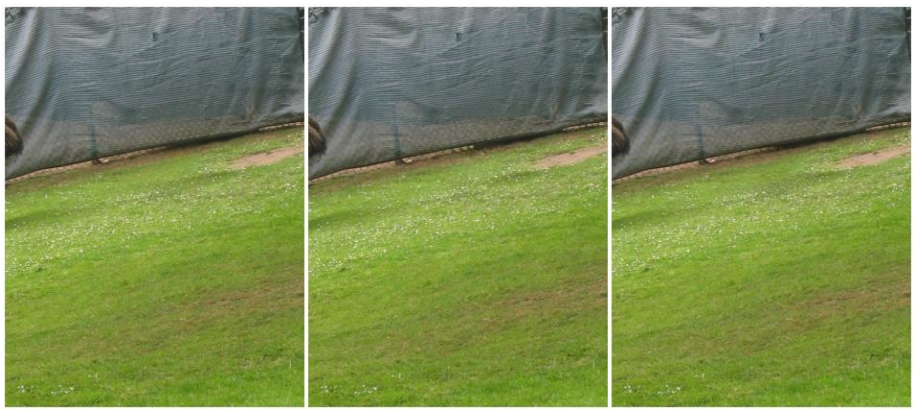

*Figure 17 Inpainted image*

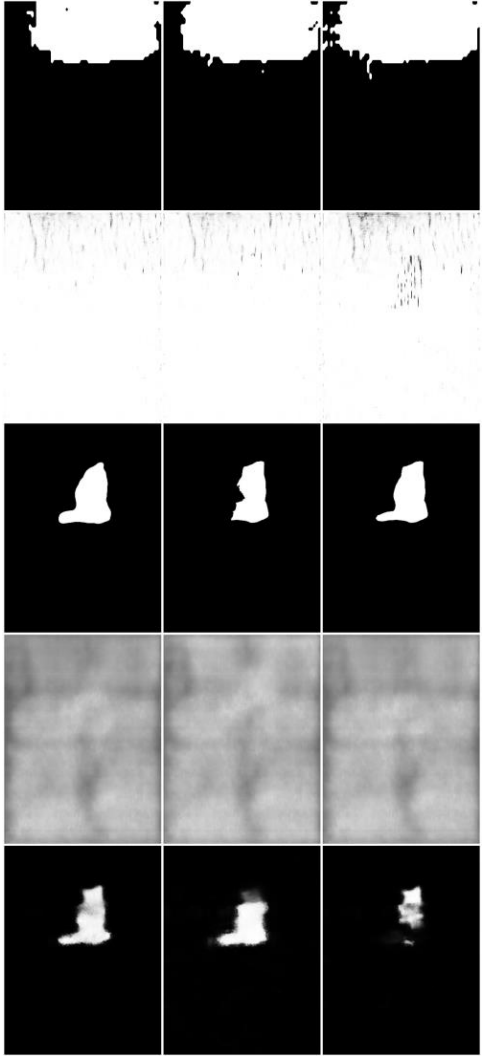

Figure 18 Detection results